\documentclass[letterpaper, 10 pt, conference]{ieeeconf}
\IEEEoverridecommandlockouts
\usepackage{cite}
\usepackage{url}

\usepackage{amsmath,amssymb,amsfonts}
\usepackage{algorithm}
\usepackage[noend]{algpseudocode}
\usepackage{graphicx}
\usepackage{textcomp}
\usepackage[table,xcdraw]{xcolor}
\usepackage{multirow}
\def\BibTeX{{\rm B\kern-.05em{\sc i\kern-.025em b}\kern-.08em
    T\kern-.1667em\lower.7ex\hbox{E}\kern-.125emX}}

\title{\LARGE \bf{Mapping-Guided Task Discovery and Allocation for Cooperative Robotic Inspection of Underwater Structures}
}

\author{Marina Ruediger$^{1}$ and Ashis G. Banerjee$^{2}$
\thanks{*This work was supported in part by the ONR grant \# N00014-21-1-2075.}
\thanks{$^{1}$Marina Ruediger is with Department of Mechanical Engineering, University of Washington, Seattle, WA 98195, USA
{\tt\small marina77@uw.edu}}%
\thanks{$^{2}$Ashis G. Banerjee is with the Department of Industrial \& Systems Engineering and Department of Mechanical Engineering, University of Washington, Seattle, WA 98195, USA
{\tt\small ashisb@uw.edu}}%
}

\begin{document}


\maketitle
\thispagestyle{empty}
\pagestyle{empty}

\begin{abstract}
Task generation for underwater multi-robot inspections without prior knowledge of existing geometry can be achieved and optimized through examination of simultaneous localization and mapping (SLAM) data. 
By considering hardware parameters and environmental conditions, a set of tasks is generated from SLAM meshes and optimized through expected keypoint scores and distance-based pruning. In-water tests are used to demonstrate the effectiveness of the algorithm and determine the appropriate parameters. 
These results are compared to simulated Voronoi partitions and boustrophedon patterns for inspection coverage on a model of the test environment. 
The key benefits of the presented task discovery method include adaptability to unexpected geometry and distributions that maintain coverage while focusing on areas more likely to present defects or damage. 
\end{abstract}


\section{Introduction}
Autonomous inspections of hulls and interior spaces are important in assessing the safety and readiness of seagoing vessels and are used to detect corrosion, biofouling, cracks, coating failures, and other damages. 
any efforts toward automating underwater inspections, but to our knowledge these are single robot systems. 
A summary of the existing tools and methods for 
underwater inspections can be found in \cite{insp-survey, 10753518}. However, many challenges persist due to high operational costs (expensive, highly customized platforms); 
huge scales of operations (detecting small defects in large vessels while operating in the open waters); and lack of system robustness and resilience (robot hardware and autonomous decision-making failures).
Consequently, we consider a low-cost cooperative multi-robot system, and present a SLAM-based task discovery and allocation algorithm to
generate the inspection tasks from environment meshes without prior geometric knowledge. This approach aims to address key underwater challenges: recovery from the loss or addition of perceptual and/or motor capabilities of robots; extremely slow and unreliable communications; and limited visibility due to turbidity and sparse features that complicate localization and mapping. 

Our method is based on a decentralized multi-robot task allocation framework that uses minimal message sizes and intermittent communications \cite{oceans}. Here, we extend this framework with an adaptive task discovery method that focuses the inspection effort on geometrically interesting areas identified during SLAM mesh generation. Specifically, we contribute a novel algorithm that automatically discovers inspection tasks from SLAM-generated meshes for heterogeneous multi-robot teams performing visual inspections of hulls or fluid-filled interior tanks. The inspection consists of visiting locations defined by position and orientation and capturing color images. Selected locations are referred to as tasks, while potential locations during discovery are candidate inspection points (CIPs). 

Section 2 outlines relevant background information on underwater SLAM and task generation algorithms. 
Section 3 details our task discovery algorithm - \underline{Map}ping-guided \underline{T}asks for \underline{I}nspection: \underline{D}iscovery and \underline{AL}location (Map-TIDAL), including the task discovery process and integration with task allocation. 
Section 4 presents the results of our in-water testing and comparisons with other task generation methods. 
In Section 5, we discuss the advantages and disadvantages, followed by our conclusions in Section 6. 

\section{Background}
\subsection{Underwater SLAM}
Many advances have been made in the field of underwater SLAM, as summarized in \cite{slam-survey,heshmat2025underwater}. Due to the extremity of the underwater environment, many typical sensors used in aerial or ground SLAM solutions are unavailable or have limitations. Vision-based sensors that work well include both monocular \cite{orb-mono} and stereo cameras, but care must be taken with shifts in color, water clarity \cite{snow-removal}, and image distortion \cite{camera-calibration}. Sonar systems are widely used for range finding-based SLAM \cite{sonar-only,sonar-slam,10682261}, as acoustic waves have significantly lower attenuation rates in water and are not affected by changes in illumination conditions. However, they are often expensive. As an alternative, LiDAR systems have been used, although they work at different frequencies of light than their aerial counterparts \cite{uwlidar}. 

Internal state-based sensors are often fused with external environmental sensors to overcome external sensing challenges. Doppler velocity logs (DVLs) use acoustic pulses to calculate velocity, compasses provide orientation data, inertial measurement units (IMUs) measure both linear and angular acceleration, and depth sensors calculate depth based on water pressure \cite{slam-survey}. Often, multiple sensors are fused together to provide a robust solution. For example, \cite{10494693,sonar-visual} combine visual, inertial, and sonar data; \cite{vis-iner-depth} uses visual, inertial and depth data; \cite{10938346} couples DVL, visual, and inertial data; \cite{rahman2022svin2} fuses sonar, visual, inertial, and water-pressure data; and \cite{xu2025tank} presents a new dataset along with a solution encompassing stereo camera, IMU, DVL, and pressure sensor data.

\subsection{Inspection Coverage}
Multi-robot (or more generally multi-agent) coverage problems appear in a wide variety of domains, including inspection and surveillance, search and rescue, agriculture, manufacturing, traffic management, law enforcement, and more. Although many solutions have been proposed, 
typically, the coverage problem is addressed either through a detailed consideration of the choice of agents' viewpoints over a complex but well-known geometry or through the partitioning of a simple bounded region for a set of agents with known capabilities.

For example, measurement uncertainties based on optical parameters are used to generate a set of viewpoints in \cite{meas-uncertainty}. An enhanced rapidly-exploring random tree is then used to compute the optimal subset of viewpoints that maintains full coverage and visibility of all the measurement points. A semiautomatic view planner is presented in \cite{insp-views}, which uses 3D triangulation between a camera and a laser to actively calculate the viewpoints from an existing mesh model. 
A multi-robot boustrophedon coverage method for unknown 2D environments using line-of-sight communications among the explorer robots is developed in \cite{multi-sweeps}. This method relies on cell splitting around obstacles to enable coverage using a heterogeneous system of a team of coverer robots and up to two teams of explorer robots. A sensor-based coverage path planning method that uses a generalized Voronoi diagram-based graph to model the energy capacity of multiple robots is presented in \cite{energy-cpp}.

\section{Method}
\subsection{Task Allocation}
Map-TIDAL employs a decentralized task allocation framework based on the consensus-based bundle algorithm (CBBA) with domain-specific modifications for underwater multi-robot inspection
\cite{oceans}. The task allocation system operates through a distributed auction mechanism, where each robot maintains autonomous decision-making capabilities while coordinating with the team through minimal message exchanges — a critical requirement for bandwidth-constrained underwater communication environments. 
\begin{figure}[htbp]
\centering
\includegraphics[width=1.0\linewidth]{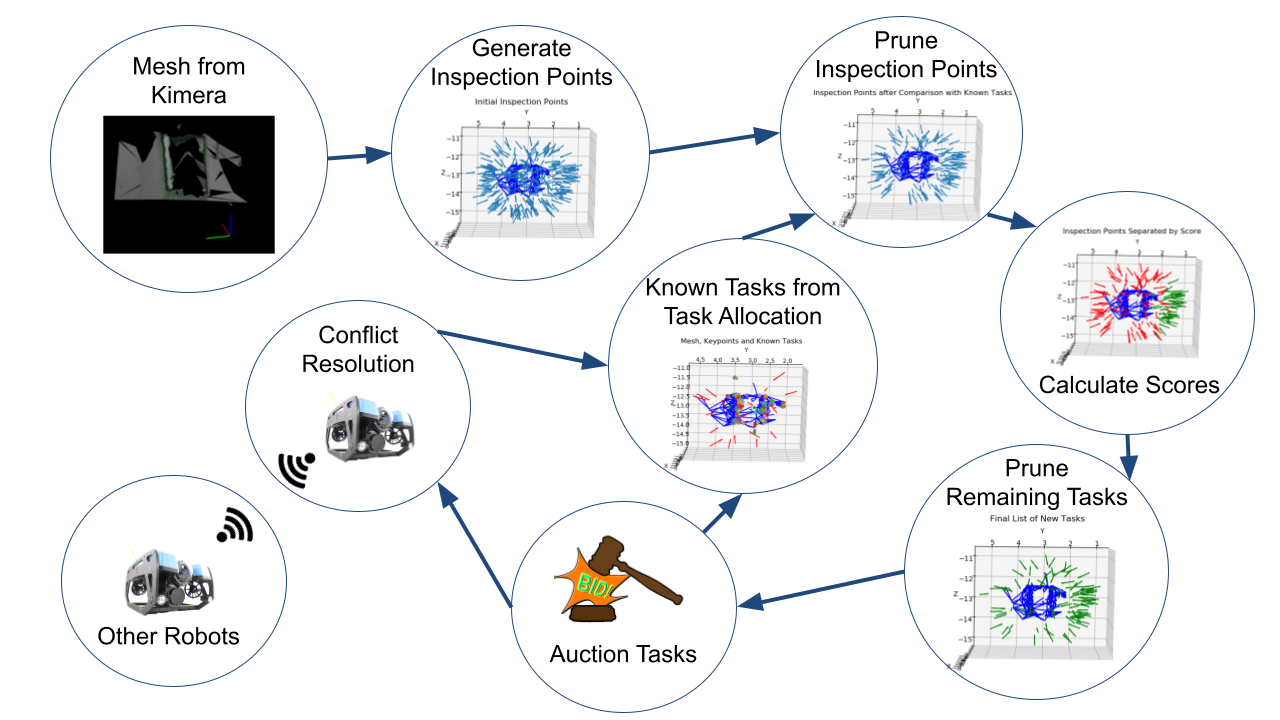}
\caption{Flowchart of the task discovery and task allocation processes.}
\label{fig:pf}
\end{figure}

Each robot executes a local task manager responsible for maintaining: (1) known task inventories, (2) bid histories from neighboring robots, (3) communication timing logs, and (4) conflict resolution protocols. The auction process utilizes a reward function based on path optimization, where the reward for incorporating a given task into a robot's bundle is calculated as the difference in total path distance when the task is optimally inserted versus the original path configuration.

The bidding mechanism employs distance-based metrics, where each robot's bid represents the cumulative distance to reach that task location. This approach ensures convergence properties while providing computationally efficient effort estimation for task completion. Upon completion of the local auction phase, bid information is disseminated to neighboring robots for decentralized conflict resolution following the consensus protocols established in \cite{cbba}.

The integration of task allocation with SLAM and task discovery processes in Map-TIDAL enables fully autonomous operation, where newly discovered tasks are seamlessly incorporated into the existing allocation framework and communicated efficiently throughout the robot network. Figure \ref{fig:pf} shows the integration between the task allocation and task discovery steps in the pipeline from the SLAM keyframe mesh through the communication and conflict resolution stages. The consensus-based task sharing inherently creates a distributed understanding of spatial coverage across the inspection area. This shared awareness of existing task locations enables intelligent mesh analysis during task discovery, where newly generated candidate inspection points can be evaluated not only for their current viewpoint coverage but also for their contribution to overall coverage gaps identified through the collective task inventory.

\subsection{Task Discovery}
In this work, we use Kimera-Multi \cite{kimera,9686955} to fuse IMU, stereo vision, and DVL data. While originally intended for aerial and ground environments, Kimera-Multi is able to handle underwater sensor fusion. With properly calibrated cameras, Kimera-Multi calculates location and generates a lightweight 3D mesh 
from very sparse keypoint data using structural regularities \cite{meshing}.
This mesh is ideal for underwater inspection 
since visual features underwater are typically clustered only in areas with strong visual differences, 
such as patches of barnacles or algae, a scratch or rust damaged area, or a piece of complex geometry like a weld line or sea chest. These visual features are the exact things that need the most attention during inspections. 
In areas with large, smooth sections with minimal features, 
the robots will rely on the DVL and IMU for localization and the mesh will be less dense and will generate fewer tasks.

There are four main steps in the task discovery part of the Map-TIDAL process which begins whenever a keyframe mesh ($\mathcal{M}$) is produced by Kimera.

\begin{figure}[htbp]
\centering
\includegraphics[width=1.0\linewidth]{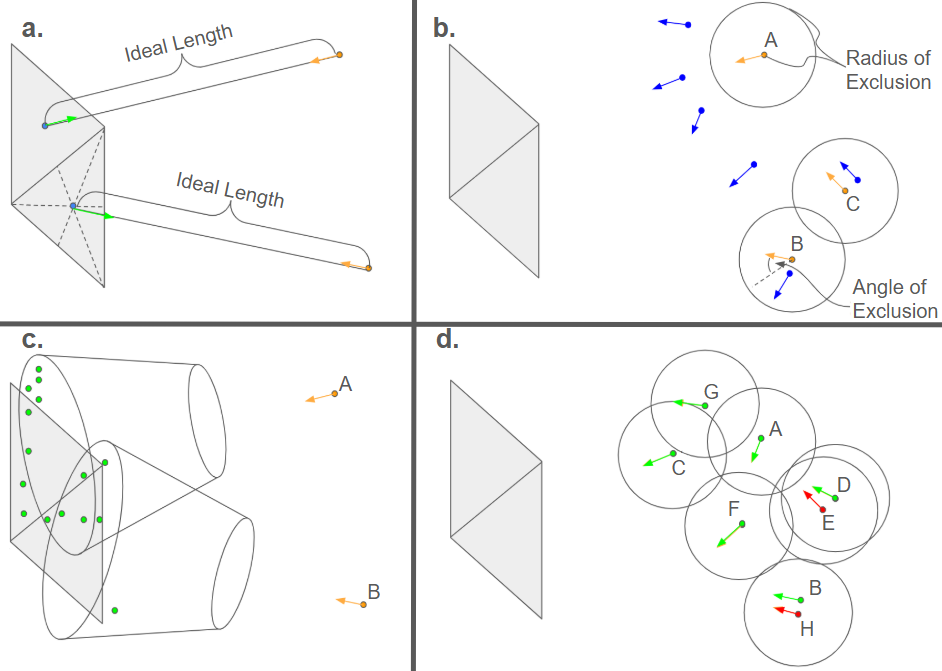}
\caption{\textbf{a.} Inspection point (orange) generation from mesh (grey) using the centroid (blue) and normal (green). \textbf{b.} Comparison of inspection points (orange) to existing tasks (blue), point A has no tasks inside the radius of exclusion, point B has a task in the radius of inclusion, but the angle is outside of the angle of exclusion, point C has a task that is within both the radius and angle of exclusion, and will not be removed. \textbf{c.} Calculation of keypoint scores based on the camera field of view. The score is the total of the keypoints (green) that fall in the field of view for the inspection point (orange) normalized by the number of keypoints. \textbf{d.} Final pruning on remaining inspection points, done in alphabetical order tasks E and H are removed, the remainder become tasks.}
\label{fig:td}
\end{figure}

\begin{enumerate}
\item \textbf{Candidate inspection point generation:} For each triangle in the mesh $\mathcal{M}$, candidate inspection points (CIPs) are generated by projecting along the surface normal by the ideal length ($L_i$), as shown in Fig.~\ref{fig:td}a. 
$L_i$ is optimized based on camera focal length and water turbidity to maximize image quality while maintaining safe standoff distance from the vessel surface.
\item \begin{enumerate} 
\item \textbf{Pruning based on existing tasks} (when $T_e \neq \emptyset$): 
Each CIP undergoes spatial filtering against existing tasks $T_e$ using dual geometric constraints: radius of exclusion ($r_{ex}$) and angle of exclusion ($\theta_{ex}$). A CIP is discarded if any existing task lies within both the spherical radius $r_{ex}$ and the angular cone $\theta_{ex}$, ensuring that retained points provide sufficiently distinct viewpoints for inspection coverage, as shown in 
Fig.~\ref{fig:td}b. If $T_e$ contains only points outside of $r_{ex}$ from any CIP, step 2b is performed instead.\\
\textbf{OR}
\item \textbf{Pruning based on self-comparison} (when $T_e = \emptyset$): When no prior tasks exist within $r_{ex}$ of any CIP, CIPs undergo self-comparison using reduced thresholds ($p_s=70\%$ of $r_{ex}$ and $\theta_{ex}$) to eliminate redundant inspection points within the candidate set, preventing over-dense task generation in highly featured areas.
\end{enumerate}
\item \textbf{Keypoint scoring:} Each remaining CIP is evaluated by projecting the camera field-of-view ($FOV$) cone and counting intersected mesh keypoints shown in Fig.~\ref{fig:td}c. CIPs scoring above the threshold ($t$) percentage of maximum possible keypoints in the local mesh region are promoted to confirmed tasks, prioritizing geometrically complex areas that correlate with potential defects. 
\item \textbf{Final spacial pruning:} 
Remaining sub-threshold CIPs undergo iterative comparison against the newly confirmed task set using the spatial exclusion criteria from Step 2a. Tasks are processed sequentially, with each addition expanding the exclusion zones for subsequent candidates, ensuring optimal spatial distribution in the final task set as shown in Fig.~\ref{fig:td}d.
\end{enumerate}

Once the new task list has been finalized, it is sent to the auction phase of Map-TIDAL, which in turn sends the new task information to the other robots in conflict resolution, and updates the master list of tasks. This full Map-TIDAL process is described in algorithm \ref{algo:task_discovery} and in Fig.~\ref{fig:pf}.

 \begin{algorithm}
 \caption{Task Discovery Process}
 \algnewcommand\algorithmicto{\textbf{in}}
 \algrenewtext{For}[2]%
   {\algorithmicfor\ #1 \algorithmicto\ #2  \algorithmicdo}
 \hspace*{\algorithmicindent} \textbf{Inputs:} Kimera mesh ($\mathcal{M}$), existing task list ($T_e$) \\
 \hspace*{\algorithmicindent} \textbf{Parameters:} $L_i$, $r_{ex}$, $\theta_{ex}$, $t$, $FOV$ \\
 \hspace*{\algorithmicindent} \textbf{Output:} New task list ($T_{new}$)
 \begin{algorithmic}[1]
   \Statex \Comment{$\hookleftarrow$ indicates appending to the end of a list}
   \Statex \Comment{$\emptyset$ indicates an empty list (ordered)}
   \State from $\mathcal{M}$ extract keypoints ($\mathcal{K}$) and polygons ($\mathcal{P}$) 
   \State $C$ $\gets$ \Call{Generate Points}{$\mathcal{P}$} \Comment{step 1}
   
   \If{$T_e$ is $\emptyset$}
     \State $C$ $\gets$ \Call{Prune}{$C$,$\emptyset$,$2$} \Comment{step 2a}
   \Else
     \State $C$ $\gets$ \Call{Prune}{$C$,$T_e$,$0$} \Comment{step 2b}
   \EndIf
   
   \State $T_{new}$ $\gets$ $\emptyset$
   \State $C_{remains}$ $\gets$ $\emptyset$
   \State $S$ $\gets$ \Call{Calculate Scores}{$C, \mathcal{K}$} \Comment{step 3}
   \For{$i$}{$length(S)$}
     \If{$S[i]$ $\geq$ $t$}
       \State $T_{new}$ $\hookleftarrow$ $C[i]$
     \Else
       \State $C_{remains}$ $\hookleftarrow$ $C[i]$
     \EndIf
   \EndFor

   \State $T_{new}$ $\hookleftarrow$ \Call{Prune}{$C_{remains}, T_{new}, 1$} \Comment{step 4}
   \State \Return{$T_{new}$}
  
 \end{algorithmic}
 \label{algo:task_discovery}
 \end{algorithm}


 \begin{algorithm}
 \caption{Pruning} 
 \algnewcommand\algorithmicto{\textbf{in}}
 \algrenewtext{For}[2]%
   {\algorithmicfor\ #1 \algorithmicto\ #2  \algorithmicdo}
 \hspace*{\algorithmicindent} \textbf{Parameters:} $L_i$, $r_{ex}$, $\theta_{ex}$, $FOV$, $p_s$ 
 \begin{algorithmic}[1]
   \Function{Prune}{$C, C_{comp}, n$}
     \State $precheck$ $\gets$ 1 \Comment{Boolean that marks if prior tasks exist within $r_{ex}$} 
     \State $C_{pruned}$ $\gets$ $\emptyset$
     \State $C_{compare}$ $\gets$ $\emptyset$
     \If{$n = 0$}   \Comment{$C$ is compared to only $C_{comp}$}
       \State $C_{compare}$ $\gets$ $C_{comp}$
     \ElsIf{$n = 1$} \Comment{$C$ is compared to $C_{comp}$ w.r.t. own tasks}
       \State $C_{pruned}$ $\gets$ $C_{comp}$
       \State $C_{compare}$ $\gets$ $C_{comp}$
     \ElsIf{$n = 2$} \Comment{$C$ is compared to itself}
       \State $C_{pruned}$ $\hookleftarrow$ $C[0]$
       \State $C_{compare}$ $\hookleftarrow$ $C[0]$
       \State $r_{ex}$ $\gets$ $r_{ex} \times p_s$ \Comment{precheck strength}
       \State $\theta_{ex}$ $\gets$ $\theta_{ex} \times p_s$
     \EndIf  
     \For{$X$}{$C$}
       \State $d$ $\gets$ $\emptyset$ \quad $\alpha$ $\gets$ $\emptyset$ 
       \State $d$ $\hookleftarrow$ $distance(X, Y)$ \textbf{for} $Y$ \textbf{in} $C_{compare}$
       \State $\alpha$ $\hookleftarrow$ $angle(X, Y)$ \textbf{for} $Y$ \textbf{in} $C_{compare}$
       \State $keep$ $\gets$ 1  \Comment{Boolean indicating whether to keep a task}     
       \For{$i$}{$0$ to $length(d)$}
         \If{$d[i]$ $<$ $r_{ex}$}
           \State $precheck$ $\gets$ 0 
           \If{$\alpha[i] < \theta_{ex}$}
             \State $keep$ $\gets$ 0
             \State $\mathbf{break}$
           \EndIf
         \EndIf
       \EndFor
       \If{$keep=1$}
         \State $C_{pruned}$ $\hookleftarrow$ $X$
         \If{$n = 1$ $\mathbf{or}$ $n = 2$}
           \State $C_{compare}$ $\hookleftarrow$ $X$
         \EndIf
       \EndIf
     \EndFor
     \If{$n = 0$ $\mathbf{and}$ $precheck = 1$}
       \State $C_{pruned}$ $\gets$ \Call{Prune}{$C$, $\emptyset$, 2}
     \EndIf
     \State \Return{$C_{pruned}$}
   \EndFunction
 \end{algorithmic}
 \label{algo:pruning}
 \end{algorithm}
 
 \begin{algorithm}
 \caption{Additional Functions} 
 \algnewcommand\algorithmicto{\textbf{in}}
 \algrenewtext{For}[2]%
   {\algorithmicfor\ #1 \algorithmicto\ #2  \algorithmicdo}
 \begin{algorithmic}[1]
   \Function{Generate Points}{$\mathcal{P}$}
     \For{$x$}{$\mathcal{P}$}
       \State $c$ $\gets$ centroid(x) 
       \State $n$ $\gets$ normal(x)
       \State $p$ $\gets$ $c$ $+$ $L_i$ $\times$ $n$
       \State $\theta$ $\gets$ $-1$ $\times$ $n$ \Comment{angle expressed as a quaternion}
       \State $C$ $\hookleftarrow$ ($p$, $\theta$)
     \EndFor
     \State \Return{$C$}
   \EndFunction
   \Statex
   \Function{Calculate Scores}{$C, \mathcal{K}$}
     \State $S$ $\gets$ $\emptyset$
     \For{$X$}{$C$}
       \State $FOV_{t}$ $\gets$ $transform(FOV, X) $
       \State $T$ $\gets$ $\sum$ $(Delaunay(FOV_{t}).find\_simplex(\mathcal{K})$ $\geq$ $0$)
       \State $S$ $\hookleftarrow$ $T/length(\mathcal{K})$
     \EndFor
     \State \Return{$S$}
   \EndFunction
 \end{algorithmic}
 \label{algo:adtl}
 \end{algorithm}

\section{Experimental Results}
\subsection{Physical Setup and Data Collection}
To validate Map-TIDAL, we collected two sets of in-situ data in an OSB test tank with saline water. We used a team of two BlueROV2 \cite{bluerov2} robots customized with an NVIDIA Jetson Xavier AGX on-board processor, a Waterliked A50 DVL (only on the robot of interest), and an OAK-D stereo depth camera. While the tether is used as a safety measure for monitoring and manual control takeover if required, all necessary processing and recording are done on the Jetson. 

\begin{figure*}[htbp]
\centering
\includegraphics[width=1.0\linewidth]{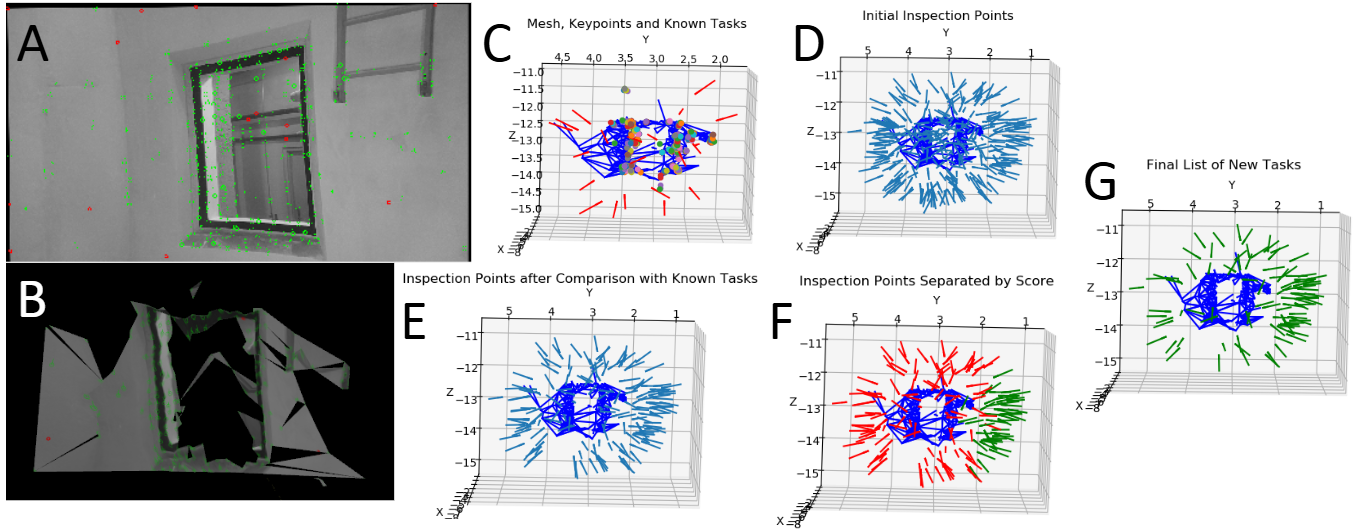}
\caption{\textbf{A.} Left camera image with tracked keypoints overlaid in green and untracked keypoints in red. \textbf{B.} Kimera mesh with A. as texture. \textbf{C.} Initial tasks shown as arrows in red, mesh in blue, keypoints as rainbow dots. \textbf{D.} The 326 CIPs generated in teal. \textbf{E.} The 165 CIPs remaining after the first pruning in teal. \textbf{F.} The 49 tasks kept from their keypoint scores in green, with the remaining 116 CIPs in red. \textbf{G.} The 98 tasks kept after the final pruning.} 
\label{fig:initial}
\end{figure*}

In the first test, a short video and IMU data with the robot manually held in a stable position were collected. This video data was then used with offline Kimera processing to determine initial feasibility and algorithm development. Figure \ref{fig:initial} shows the results of this initial algorithm development on a single mesh frame featuring 326 triangles. Initial parameters were chosen as $L_i=1.5$ m, $r_{ex}=0.8$ m, $\theta_{ex}=0.1$ rad, and $t=0.4$.
The left image with keypoints and the mesh are shown in Fig.~\ref{fig:initial}A and Fig.~\ref{fig:initial}B, showing the concentration of the keypoints on the geometrically and visually interesting places such as the ladder feet and edges of the window and frame, and the two visible edges of the corner. Using these values and an initial list of 21 tasks selected randomly from the CIPs, the number of CIPs is reduced from 326 to 98. This can be further reduced with appropriate parameter tuning. 

In the second test, the robot was allowed to rise slowly a distance of about 0.5 m, and the first 7 keyframes were analyzed to find the ideal parameters for this environment. In pretest observations, with high water clarity and good, indoor lighting at the saltwater tank, we found the OAK-D provided the most accurate point clouds when it was positioned approximately 1 meter from the surface it observed, this was set as the $L_i$ value. 

Parameters $r_{ex}$, $\theta_{ex}$, and $t$ were varied separately with the goals of minimal uninspected area, minimal overcoverage, and between 15-25 tasks. 
We observed that $\theta_{ex}$ produces the number of tasks in that range for values between 0.25 and 0.45 rad, $r_{ex}$ produces tasks in that range between 1.5 and 2.0 m, with the rate of adding additional tasks higher with lower values of $r_{ex}$ and $\theta_{ex}$. The threshold has little effect on the rate of adding tasks, but a large effect on the initial number of tasks, with higher thresholds producing fewer tasks overall, and very little change in the number of tasks produced for any threshold variance over 80\%. 

\subsection{Analysis}

\begin{figure}[htbp]
\centering
\includegraphics[width=1.0\linewidth]{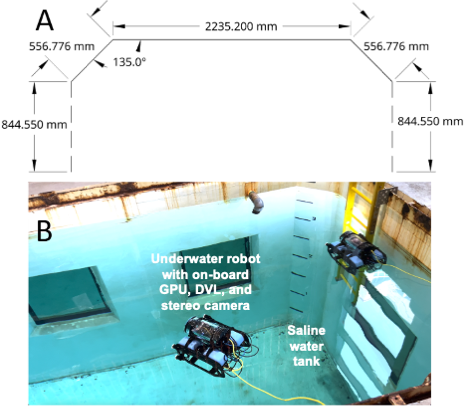}
\caption{\textbf{A.} Top view of the test tank with dimensions. Solid lines show geometry used to generate inspection points for Voronoi and sweeping pattern methods, dashed lines show the edges of the voxelized region used to evaluate method effectiveness. \textbf{B.} Test tank with two modified BlueROV2s with the main robot at the center.} 
\label{fig:tank_geom}
\end{figure}

To analyze the effective coverage of the new task locations, we modeled the region of the test tank used during testing. This region includes a flat section with an inset window, two angled sections, and a short part of the longer sides of the tank, as shown in Fig.~\ref{fig:tank_geom}. The bottom of the tank and the surface of the water are not included in the model. The model is voxelized using Open3D libraries. 

For each point in the discovered task sets, the $FOV$ cone is transformed, and any voxels within the cone are shaded red. If a voxel is seen from multiple positions, the shade of red is made brighter with each point that views it. By analyzing the distribution of the number of times each voxel is seen, we can compare the effectiveness of the variations of each parameter within their effective ranges. 

Table~\ref{tbl:params} shows the average number of views per voxel, the number of tasks generated, and the percentage of voxels with zero views for various values of $r_{ex}$, $\theta_{ex}$ and $t$. Ideal values with less than 5\% uninspected are shown in green, and for these values, between 15 and 25 tasks, and an average of less than 2.5 views per voxel are also shown in green. Yellow values indicate values outside of that range, but under 30 tasks or with an average lower than 3.5, while red values indicate averages over 3.5. The best coverage is achieved at $r_{ex}=1.7\ m$, $\theta_{ex}=0.25\ rad$, and $t=75\%$ but other values also show acceptable results. In general, lower $r_{ex}$ and $\theta_{ex}$ values lead to higher numbers of tasks with better coverage, but a higher average coverage. 

\begin{table}[t]
\caption{Average Views per Voxel, Number of Tasks, and Percent Uninspected}
\centering
\begin{tabular}{|c|c|cccccc|}
\hline
                       &                                & \multicolumn{6}{c|}{$\theta_{ex}$ (rad)}                                                                                                                                                                                                                                                                                               \\ \cline{3-8} 
                       &                                & \multicolumn{3}{c|}{0.25}                                                                                                                                                                            & \multicolumn{3}{c|}{0.3}                                                                                                        \\ \cline{3-8} 
\multirow{-3}{*}{$t$}  & \multirow{-3}{*}{$r_{ex}$ (m)} & \multicolumn{1}{c|}{avg}                         & \multicolumn{1}{c|}{num}                                               & \multicolumn{1}{c|}{\%}                                                  & \multicolumn{1}{c|}{avg}                         & \multicolumn{1}{c|}{num}                       & \%                          \\ \hline
                       & 1.5                            & \multicolumn{1}{c|}{{\color[HTML]{FE0000} 4.26}} & \multicolumn{1}{c|}{{\color[HTML]{FFC702} 27}}                         & \multicolumn{1}{c|}{\cellcolor[HTML]{FFFFFF}{\color[HTML]{32CB00} 1.49}} & \multicolumn{1}{c|}{{\color[HTML]{FE0000} 3.97}} & \multicolumn{1}{c|}{{\color[HTML]{FFC702} 25}} & {\color[HTML]{32CB00} 3.71} \\ \cline{2-8} 
                       & 1.6                            & \multicolumn{1}{c|}{3.63}                        & \multicolumn{1}{c|}{24}                                                & \multicolumn{1}{c|}{7.60}                                                & \multicolumn{1}{c|}{3.19}                        & \multicolumn{1}{c|}{21}                        & 8.06                        \\ \cline{2-8} 
                       & 1.7                            & \multicolumn{1}{c|}{{\color[HTML]{FFC702} 2.85}} & \multicolumn{1}{c|}{{\color[HTML]{32CB00} 22}}                         & \multicolumn{1}{c|}{\cellcolor[HTML]{FFFFFF}{\color[HTML]{32CB00} 4.28}} & \multicolumn{1}{c|}{2.30}                        & \multicolumn{1}{c|}{18}                        & 21.16                       \\ \cline{2-8} 
\multirow{-4}{*}{70\%} & 1.8                            & \multicolumn{1}{c|}{2.71}                        & \multicolumn{1}{c|}{19}                                                & \multicolumn{1}{c|}{6.18}                                                & \multicolumn{1}{c|}{2.79}                        & \multicolumn{1}{c|}{19}                        & 5.77                        \\ \hline
                       & 1.5                            & \multicolumn{1}{c|}{{\color[HTML]{FE0000} 4.05}} & \multicolumn{1}{c|}{\cellcolor[HTML]{FFFFFF}{\color[HTML]{FFC702} 26}} & \multicolumn{1}{c|}{{\color[HTML]{32CB00} 3.05}}                         & \multicolumn{1}{c|}{3.50}                        & \multicolumn{1}{c|}{23}                        & 5.69                        \\ \cline{2-8} 
                       & 1.6                            & \multicolumn{1}{c|}{3.12}                        & \multicolumn{1}{c|}{22}                                                & \multicolumn{1}{c|}{11.73}                                               & \multicolumn{1}{c|}{2.83}                        & \multicolumn{1}{c|}{20}                        & 9.53                        \\ \cline{2-8} 
                       & 1.7                            & \multicolumn{1}{c|}{{\color[HTML]{32CB00} 2.56}} & \multicolumn{1}{c|}{{\color[HTML]{32CB00} 20}}                         & \multicolumn{1}{c|}{\cellcolor[HTML]{FFFFFF}{\color[HTML]{32CB00} 3.81}} & \multicolumn{1}{c|}{2.38}                        & \multicolumn{1}{c|}{18}                        & 6.79                        \\ \cline{2-8} 
\multirow{-4}{*}{75\%} & 1.8                            & \multicolumn{1}{c|}{2.48}                        & \multicolumn{1}{c|}{18}                                                & \multicolumn{1}{c|}{6.84}                                                & \multicolumn{1}{c|}{2.55}                        & \multicolumn{1}{c|}{18}                        & 6.42                        \\ \hline
                       & 1.5                            & \multicolumn{1}{c|}{{\color[HTML]{FFCB2F} 3.26}} & \multicolumn{1}{c|}{{\color[HTML]{32CB00} 23}}                         & \multicolumn{1}{c|}{{\color[HTML]{32CB00} 4.06}}                         & \multicolumn{1}{c|}{{\color[HTML]{FFC702} 3.06}} & \multicolumn{1}{c|}{{\color[HTML]{32CB00} 21}} & {\color[HTML]{32CB00} 4.40} \\ \cline{2-8} 
                       & 1.6                            & \multicolumn{1}{c|}{2.62}                        & \multicolumn{1}{c|}{22}                                                & \multicolumn{1}{c|}{11.90}                                               & \multicolumn{1}{c|}{2.31}                        & \multicolumn{1}{c|}{19}                        & 8.07                        \\ \cline{2-8} 
                       & 1.7                            & \multicolumn{1}{c|}{{\color[HTML]{FE0000} 4.06}} & \multicolumn{1}{c|}{{\color[HTML]{FFC702} 25}}                         & \multicolumn{1}{c|}{{\color[HTML]{32CB00} 3.71}}                         & \multicolumn{1}{c|}{3.15}                        & \multicolumn{1}{c|}{21}                        & 5.13                        \\ \cline{2-8} 
\multirow{-4}{*}{80\%} & 1.8                            & \multicolumn{1}{c|}{3.78}                        & \multicolumn{1}{c|}{23}                                                & \multicolumn{1}{c|}{6.35}                                                & \multicolumn{1}{c|}{3.09}                        & \multicolumn{1}{c|}{20}                        & 6.45                        \\ \hline
\end{tabular}
\begin{tabular}{p{3.3in}} 
Avg indicates the average views per voxel, num the number of tasks generated with that set of parameters, and \% the percentage of voxels that receive no views. Green values indicate optimal performance, yellow values acceptable performance, and red values indicate significant overcoverage. 
\end{tabular}
\label{tbl:params}
\end{table}

\subsection{Comparison Metrics}
Metrics on inspection quality such as the coverage percent and averages can be easily compared, but finding a metric that also shows the distribution of overcoverage can be a challenge. Some overcoverage can be desired, especially around complex geometry to ensure that all angles are viewed to reveal all defects, but too much overcoverage is a sign that there are more inspection points than needed- leading to wasted time and energy. But what is the ideal balance that ensures coverage without too much overcoverage? While many options are present, we have chosen to examine the distribution of number of views per voxel compared to the gamma distribution. While this function is typically used for wait time variables, the inherent 0 minimum, left skew, and exponentially decreasing tail make the gamma distribution good for our purposes as well. The gamma distribution has the probability density function in equation~\ref{eq:gamma-pdf}.

\begin{equation}
\centering
f(x;\alpha,\beta)=\frac{1}{\Gamma(\alpha)\beta^{\alpha}}x^{\alpha-1}e^{-x/\beta}
\label{eq:gamma-pdf}
\end{equation}

By choosing the shape parameter $\alpha$ and the scale parameter $\beta$, we can modify the shape of the distribution to fit our desired inspection requirements- that there be a minimal number of voxels inspected 0 times, that the peak value be between 1 and 2, and that the number of voxels with large amounts of overcoverage be low. The peak value of the gamma distribution is at $\beta(\alpha-1)$, so if we set a desired peak value, either parameter can be used to tune the whole distribution. Using 2 as the peak value and setting $\alpha=4$ gives $\beta=\frac{2}{3}$, and checking with the cumulative density function, this choice of parameters gives less than 1\% of voxels being inspected less than 0.5 times. Since the gamma distribution is continuous while the number of views is discrete integer values, when discretizing data from the gamma distribution, this would be the cut-off for voxels that would be inspected 0 times using integer centered bins. 

\subsection{Comparisons}

\begin{figure*}[htbp]
\centering
\includegraphics[width=0.85\linewidth]{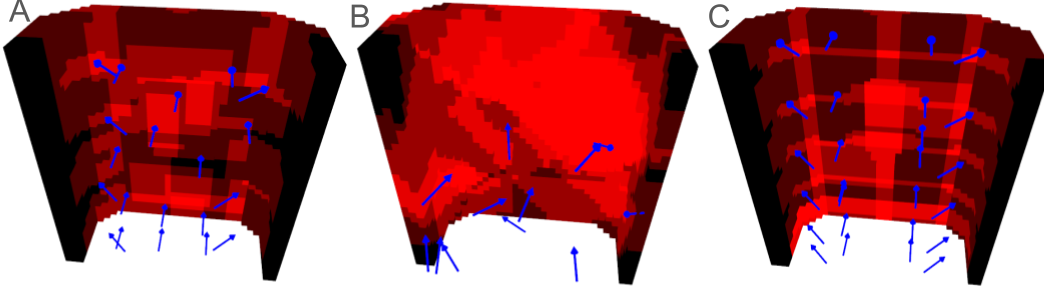}
\caption{Simulated coverage analysis for \textbf{A} Voronoi, \textbf{B} Map-TIDAL, and \textbf{C} sweeping patterns with task locations shown in blue and voxels colored more red with each time they are viewed.}
\label{fig:voxels}
\end{figure*}

\begin{figure}[htbp]
\centering
\includegraphics[width=0.88\linewidth]{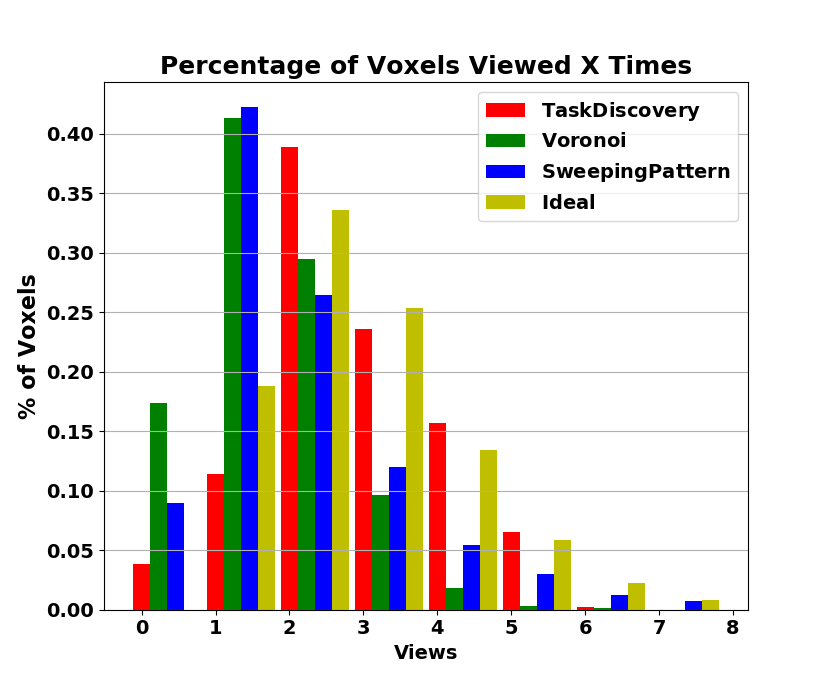}
\caption{Histogram showing normalized views per voxel for different task generation methods compared with the discretized ideal gamma distribution.}
\label{fig:histo}
\end{figure}

Some understanding of other prevalent area coverage methods is required to compare them to the performance of MAP-Tidal. Voronoi diagrams are made by dividing a space into regions that contain all the points closest to a point within a set of points. A survey of Voronoi methods for robotic coverage is provided in \cite{voronoi-survey}. The generation of the points used to generate the Voronoi diagram can be generated by using robot start positions, random generation based on various distributions, or other various methods. The Voronoi partitions can be generated with a constant cost function or on a distributed cost function over the region based on previous knowledge of the environment. From the starting positions, the diagram can be optimized using Lloyd's method, which moves each point toward the centroid of the corresponding Voronoi partition to recompute the new partitions. For our comparisons, we use WebSVG's Voronoi editor \cite{voronoi}, which uses a seed generation method that controls the spread and regularity and allows for weighted distributions and wall avoidance. 

Boustrophedon patterns are based on sweeping linearly back and forth over a space. In \cite{lawnmowers}, the optimization of sweeping patterns occurs over a space that can be represented by a 2D space with holes as no-fly zones. Bahnemann et al. develop a tool to calculate optimized paths using a generalized traveling salesman problem framework that can be modified during runtime as a user adds additional no-fly zones. While this implementation was originally demonstrated using flying robots searching a hillside for mines with a set altitude offset, it can be applied to other situations where a 3D space can be represented by a surface with an offset robot, such as our inspection problem. 

For both the Voronoi and sweeping patterns, we use a projection of the rectangular surface onto the wall of the tank with a distance offset of 1 m and an angle perpendicular to the surface of the tank. A top-down section of the portion of the tank considered is shown in Fig.~\ref{fig:tank_geom}. Similar to the evaluation of coverage used for parameter adjustment, the same tank model is voxelized and the same parameters can be used to evaluate the effectiveness of the coverage of these methods. 

Fig.~\ref{fig:voxels} shows the best version of each generation method and Fig.~\ref{fig:histo} shows a histogram chart with the views per voxel overlaid with the ideal gamma based distribution generated in part IV.C. The Boustrophedon pattern was generated using two interpolation points between sections longer than 1 m, with a lateral overlap of 0.1 m. The Voronoi partition is generated with 20 points including wall avoidance. Map-TIDAL used $r_{ex}=1.7$ m, $\theta_{ex}=0.25$ rad, and $t=75\%$. 

From the histogram in Fig.~\ref{fig:histo}, it can be seen that Map-TIDAL achieves 3.81\% uninspected voxels with an average of 2.56 views per voxel in 20 tasks as compared to the Voronoi (17.32\% uninspected, 1.39 average views, 20 tasks) and boustrophedon (8.94\% uninspected, 1.80 average views, 30 tasks). The average of absolute differences between Map-TIDAL and the ideal is 0.029, while the Voronoi (0.100) and boustrophedon (0.081) display larger differences.  

\subsection{Additional Experimental Verification}

\begin{figure*}[htbp]
\centering
\includegraphics[width=0.87\linewidth]{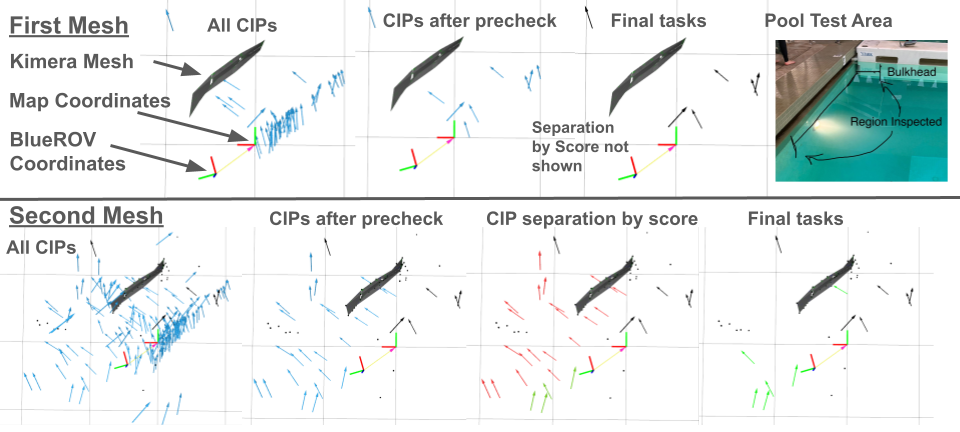}
\caption{Task discovery with steps shown over the first two mesh frames in pool testing. CIPs are colored blue initially. In the case of the first mesh, the final tasks are shown in black, which remain when they are used in the second mesh as the existing tasks. For the second mesh, during the score separation step, ones that are selected are green, while the ones subject to final pruning are red, and the final selection is shown in light green. These light green tasks are then combined with the existing list of tasks for the next mesh frame.} 
\label{fig:pooltest}
\end{figure*}

The third test was conducted in a fresh water pool with no additional parameter tuning, as the pool visibility was similar to that in the OSB test tank. This test involved more movement of the robot, with a surface not directly perpendicular to the robot, similar to what it would see if moving along a hull at an angle. 

Figure~\ref{fig:pooltest} shows the results of the first and second mesh frames in one instance of pool based testing. Based on observations made during testing, this set of tasks appears well distributed over the environment, though it can be noted that the Kimera mesh visualization is not complete. The mesh includes all triangles in the mesh while the visualization includes only ones over a certain size. It can also be noted in later mesh frames (not shown) that the reflection of the wall onto the surface of the water causes additional mesh irregularities, indicating that additional visual processing should be developed to handle regions near the surface. 

Overall, this third test was very successful, demonstrating the functionality of the MAP-Tidal algorithms in different environments with different geometric complexities. 

\section{Discussion}
Qualitative analysis of the areas with the brightest reds in the task discovery method from Fig.~\ref{fig:voxels}B, shows clear concentration of views on the center and upper right side of the tank. This corresponds with the position of the safety ladder during the testing, so it is expected that that region should be viewed from more angles due to the concentration of keypoints in that area. Both the sweeping and Voronoi patterns lack this concentration in the provided examples, but with modifications to weighting parameters both methods have the potential to incorporate weighted distributions to task generation. 

The main limitation of the Voronoi and sweeping patterns for the underwater inspection case is that they require a model before they can generate their points. The sweeping pattern tool is able to recalculate quickly on the fly when a user adjusts the model of the space, but still requires user input, preventing the system from being fully autonomous. The Voronoi patterns also calculate quickly, but require the geometric limits of the space before starting. Changing the environment for either set would overwrite the existing tasks with a new set, which would be challenging, since underwater communication is slow and unreliable. 
This limits the effectiveness of both methods in dynamic environments with minimal a priori knowledge of the geometry. 

The proposed Map-TIDAL method is also not without limitations. Since task discovery is based on the meshes generated during SLAM, it does not require prior knowledge of the geometry to be inspected, but the quality of the generated tasks will be heavily dependent on the quality of the SLAM meshes. In murky water, additional sensing, such as the DVL, is necessary to maintain localization when no objects are visible, since any errors in localization will result in corresponding errors in task positions. In addition, with only the geometry seen in SLAM by the individual robot available during task discovery, there is the possibility that the generated tasks may be placed in areas the robots cannot access. 

\section{Conclusion}

Future works will include investigation into collision avoidance during SLAM based task discovery and allocation. Another potential for future work exists in the multi-modal sensor fusion domain, where additional fusion with side scan sonar data could augment localization and map generation abilities. Side scan sonar data would also provide more accuracy for Map-TIDAL's task discovery as well as providing mesh accuracy verification. While the Map-TIDAL method currently focuses on visual inspection data, the algorithms could be extended by developing adaptations for other inspection sensors such as ultrasonic thickness testing or for the use of additional tools such as brushes, scrapers, or probes. Map-Tidal is designed for underwater use, but has potential for use in challenging aerial environments with limited communications and visual range if adapted for a system of flying drones. 

Map-TIDAL shows that task discovery using a SLAM generated environment mesh  can be achieved by generating candidate inspection points and pruning them based on their keypoint scores and proximity to other tasks. These tasks can then be efficiently distributed to a multi-robot system through a decentralized auctioning and conflict resolution process. As new geometry is discovered, new tasks are generated and distributed to the system, removing the need for models of the objects to be inspected and focusing the inspection on areas with large numbers of visually distinct features. With proper parameter adjustment and accurate SLAM, Map-TIDAL produces task locations that provide excellent coverage with appropriately low levels of overinspection.

\section*{Acknowledgment}

We would like to thank Wade Kempf, Tyler Paine, Matthew Craw, and Martin Renken for their mentorship and support of our research.  We acknowledge the use of NIPRGPT in light-editing of our text.

\bibliographystyle{ieeetr}
\bibliography{ref}

\end{document}